\documentclass{article}
\usepackage{spconf,amsmath,graphicx,algorithm,algpseudocode}

\title{ITERATIVE LOW-RANK APPROXIMATION FOR CNN COMPRESSION}
\name{Maksym Kholiavchenko}
\address{Computer Science and Engineering, Innopolis University, Russia}
\begin{document}
%\ninept
%
\maketitle
\begin{abstract}
Deep convolutional neural networks contain tens of millions of parameters, making them impossible to work efficiently on embedded devices. We propose iterative approach of applying low-rank approximation to compress deep convolutional neural networks. Since classification and object detection are the most favored tasks for embedded devices, we demonstrate the effectiveness of our approach by compressing \mbox{AlexNet \cite{krizhevsky2012imagenet}}, VGG-16 \cite{simonyan2014very}, YOLOv2 \cite{redmon2016yolo9000} and Tiny YOLO networks. Our results show the superiority of the proposed method compared to non-repetitive ones. We demonstrate higher compression ratio providing less accuracy loss.
\end{abstract}
\begin{keywords}
Low-Rank Approximation, Neural Network Compression, Object Detection Optimization, Higher-order singular value decomposition
\end{keywords}
\section{Introduction}
\label{sec:introduction}
Convolutional neural networks have shown their efficiency for a wide range of tasks \cite{krizhevsky2012imagenet, simonyan2014very, redmon2016yolo9000, szegedy2015going, he2016deep, ren2015faster, he2017mask, gatys2015neural}. Deep models demonstrate state-of-the-art results and achieve human-level performance \cite{taigman2014deepface, rajpurkar2017chexnet}. Thereby, the development of deeper and more complicated networks in order to achieve higher accuracy has become commonplace. But such networks contain tens of millions of parameters and cannot be efficiently deployed on embedded systems and mobile devices due to their computational and power limitations. 

Deep neural networks make use of parameter redundancy to facilitate convergence \cite{hinton2012improving}. Such redundancy can be eliminated from trained deep neural networks by applying low-rank approximation to compress weight tensors. As a result, compression reduces model size and speed up execution time. It can greatly facilitate the distribution of the model and reduce the number of computations that positively affects user experience and power consumption. Several studies have been conducted in this area \cite{lebedev2014speeding, kim2015compression}. However, despite the success in applying compression, there is the problem of accuracy loss.

This paper describes an algorithm to address this problem and also increase rate of compression. We introduce \textit{iterative low-rank approximation algorithm} for automated network compression. The algorithm consists of four repetitive steps: automated selection of extreme rank values, rank weakening, tensor decomposition according to weakened rank and fine-tuning. We found that optimal number of iterations depends on extent of rank weakening and is in the range of from 2 to 4.

\begin{figure}[tb]
\includegraphics[width=8.5cm]{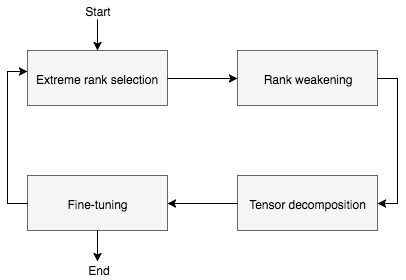}
\centering
\caption{Iterative low-rank approximation algorithm. Input: model (configuration and weights) to be compressed. Output: compressed model (configuration and weights).}
\label{fig:image1}
\end{figure}

Redundancy in deep neural networks can be considered as noise that contains a very small percentage of variance. Basically, our approach gradually perform noise reduction by repeating two phase: get rid of noise, fine-tune and introduce new noise. In the first phase, we remove a very small amount of noise at once. It allows to completely recover accuracy by performing the second phase. By repeating these phases, we can gradually compress a model and unlike other approaches, do not lose accuracy by reducing extreme amount of variance at once. It is also worth to mention that gradual approach allows to achieve higher compression ratio since removing more dimensions than allowed by found extreme rank imposes drastic accuracy drop.

Proposed algorithm can be reproduced with any deep learning framework since it does not require altering the framework itself but only altering the model configuration and model weights. Our experiments were mainly carried out using Caffe and Darknet frameworks. The results show the superiority of our iterative approach with gradual compression in comparison with non-repetitive ones.

\section{Background}
\label{sec:background}
There are several works devoted to the deep convolutional neural networks compression. Authors of \cite{han2015deep} proposed a pipeline that consists of three different methods: pruning, trained quantization and Huffman coding. They demonstrated the possibility of the significant reducing of storage requirement by combining different techniques. But our method is different because we focus not only on compression ratio, but also on the corresponding speedup and seamless integration into any framework.

Various methods based on quantization have been proposed by \cite{lin2016fixed, miyashita2016convolutional, park2017weighted}. The main goal of quantization is to reduce the number of bits required for weight storage. Our approach is also different because we compress networks by decomposing tensors and reducing rank. It is worth to mention that quantization can be used after applying our method and serve as the second method of compression. But quantization may require altering a framework and a significant speed up can be achieved only taking into account the peculiarities of the hardware.

There are several experiments on training low precision networks \cite{baldassi2015subdominant, rastegari2016xnor}. Their methods allow to use only 2 bits for weight storage but accuracy is much lower than in full precision networks and it is not a compression algorithms because such networks have to be trained from scratch.

Several approaches based on different algorithms of low-rank approximation were proposed by \cite{lebedev2014speeding, denton2014exploiting}. Authors of \cite{denton2014exploiting} have demonstrated success of applying singular value decomposition (SVD) to fully connected layers. And authors of \cite{lebedev2014speeding} found a way to decompose 4-way convolutional kernel tensor by applying canonical polyadic (CP) decomposition. But these approaches are able to compress only one or couple layers. Moreover, for each layer the rank is unique and process of rank selection have to be performed manually every time.

Another way to compress a whole network was introduced by \cite{kim2015compression}. The approach used in their work is automated. Authors combined two different decompositions to be able to compress both fully connected and convolutional layers. To compress fully connected layers they adopted approach used by \cite{denton2014exploiting} and applied SVD. For convolutional layers authors applied a Tucker decomposition \cite{tucker1966some}. Unlike \cite{lebedev2014speeding, denton2014exploiting}, authors found a way to automatically select ranks without any manual search. Ranks are determined by a global analytic solution of variational Bayesian matrix factorization (VBMF) \cite{nakajima2012perfect}. We found that the global analytic VBMF provides ranks for which it is difficult to restore initial accuracy by fine-tuning for deep networks. In our algorithm, we also use the global analytic VBMF but to select extremal ranks which will be weakened afterwards.

CP decomposition which was used by \cite{lebedev2014speeding} is a special case of a Tucker decomposition, where the core tensor is constrained to be superdiagonal. In our approach we use higher-order singular value decomposition (HOSVD) which is also a specific orthogonal Tucker decomposition. To compress fully connected layers we adopt SVD as it was proposed by \cite{denton2014exploiting}.

Our approach is different from these methods because all of them apply decomposition algorithm only one time per layer and ranks provided by the global analytic VBMF can be considered as upper bound for compression. Our algorithm is iterative and decomposition algorithm can be applied couple times for the same layer. Moreover, we can achieve higher compression rate because we do not have such boundary.

\section{Decomposition and rank selection}
\label{sec:decomposition_and_rank_selection}
Our algorithm consists of four iterative steps: automated selection of extreme rank values, rank weakening, tensor decomposition according to weakened rank and fine-tuning. In this section we describe the first three steps.

\subsection{Decomposition algorithms}
\label{ssec:decomposition_algorithms}
Convolutional neural network can be comprised of convolutional layers followed by fully connected layers. Convolutional layers is a 4-way kernel tensors and fully connected layers is a 2-way kernel tensors. In general, 2-way kernel tensor is a matrix. To compress fully connected layers, we employ singular-value decomposition (SVD). Singular value decomposition is a method of decomposing a matrix into three other matrices:

\begin{equation}
A=U × S × V^T
\end{equation}

\noindent
where, $A$ is an ($m \times n$) matrix, $U$ is an ($m \times n$) orthogonal matrix, $S$ is an ($n \times n$) diagonal matrix and $V$ is an \mbox{($n \times n$)} orthogonal matrix. After decomposition we can obtain approximated matrix:

\begin{equation}
a_{ij} \approx \sum_{k=1}^{p} × u_{ik} × s_{k} × v_{jk}^T
\end{equation}

\noindent
where, $p$ is the rank value. To compress convolutional layers, we employ higher-order singular value decomposition (HOSVD). It is a specific orthogonal \mbox{Tucker decomposition \cite{tucker1966some}} and can be regarded as a generalization of the matrix SVD. Originally, the Tucker decomposition decomposes a tensor into a core tensor multiplied by a matrix along each mode:

\begin{equation}
A = B \times_{1} C^{(1)} \times_{2} C^{(2)} \times_{3} \ldots \times_{n} C^{(n)}
\label{equation1} 
\end{equation}

\noindent
where, $A$ is the original n-way tensor, $B$ is the core tensor, $C^{(n)}$ are matrices to be multiplied along each mode and $\times_{k}$ is the k-mode product. HOSVD is a way to solve Tucker decomposition by solving each mode-k matricization of the Tucker decomposition for $C^{(k)}$.

\begin{figure}[tb]
\includegraphics[width=8.5cm]{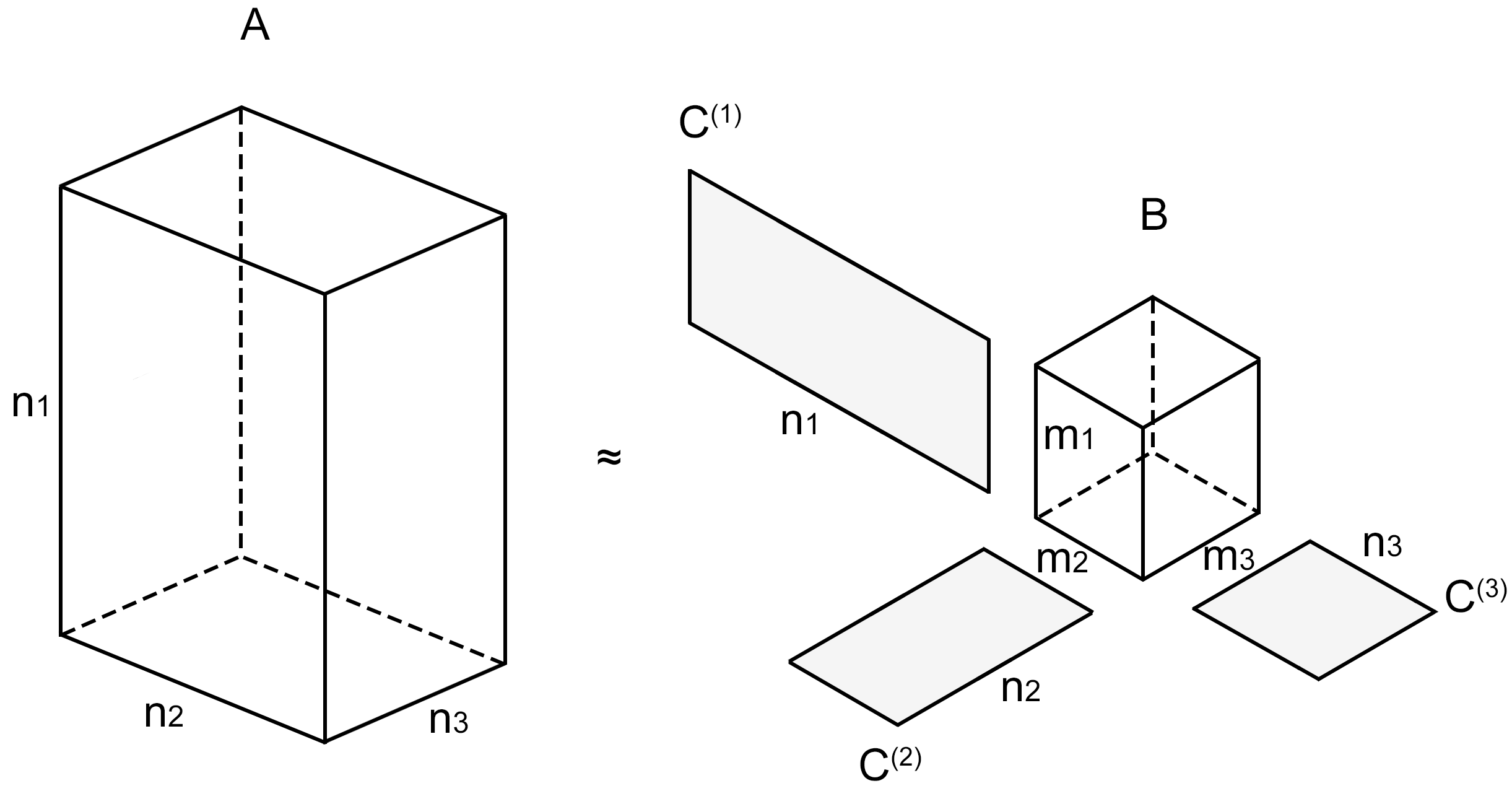}
\centering
\caption{Higher-order singular value decomposition of a 3-way tensor. $A$ the 3-way tensor. $B$ is the core 3-way tensor. $C^{(1)}$, $C^{(2)}$, $C^{(2)}$ are matrices to be multiplied along each mode of core tensor $B$.}
\label{fig:image2}
\end{figure}

\pagebreak
\noindent
Higher-order singular value decomposition relies on fact that following equation is equivalent to equation \ref{equation1}:
\begin{equation}
\resizebox{0.44 \textwidth}{!} 
{
$A_{(k)} = C^{(k)} B_{(k)} \left(C^{(n)} \otimes \cdots \otimes C^{(k+1)} \otimes C^{(k-1)} \otimes \cdots C^{(1)}\right)^T$
}
\end{equation}

\noindent
where, $X_{(k)}$ is the mode-k matricization of the tensor $B$ and $\otimes$ is the Kronecker product.

\subsection{Rank selection algorithm}
\label{ssec:rank_selection_algorithm}
For convenience, we introduce two concepts of rank: an \textit{extreme rank} and a \textit{weakened rank}. The extreme rank is the value at which almost all noise is eliminated from the tensor after decomposition. The weakened rank is the value at which a certain amount of noise is preserved in the tensor after decomposition.

Manual rank selection is a time consuming operation and violates idea of automated network compression. To automatize process of rank selection, we adopted approach proposed by \cite{kim2015compression}. The global analytic VBMF can automatically find noise variance. But unlike authors of \cite{kim2015compression}, we use global analytic solutions for variational Bayesian matrix factorization (VBMF) \cite{nakajima2012perfect} only for determining extreme rank. It is worth to mention that the global analytic VBMF gives a suboptimal solution. We consider values provided by the global analytic VBMF as upper bound for one-time compression. To determine ranks using the global analytic VBMF, we apply it only on mode-3 and mode-4 matricization of a kernel tensor because mode-1 and mode-2 are associated with spatial dimensions \cite{kim2015compression}. They are already quite small and do not have to be decomposed by HOSVD.

The weakened rank depends linearly on extreme rank and serves to preserve more noise in decomposed tensor. It facilitates fine-tuning and allows to perform decomposition more than once.

\pagebreak
\noindent
The weakened rank is defined as follows:

\begin{equation}
R_{w} = 
\begin{cases} 
R_{i} - k × (R_{i} - R_{e}), & R_{i} > 20 \\
R_{i}, & otherwise \\
\end{cases}
\end{equation}

\noindent
where, $R_{w}$ is the weakened rank, $R_{i}$ is an initial rank, $R_{e}$ is the extreme rank and $k$ is a weakening factor. In this formula, $k$ is a hyperparameter such that: $0 < k < 1$. Our experiments show that optimal value for $k$ is in the range: $0.5 \leq k\leq 0.7$. If initial rank is less than $21$, our algorithm considers such kernels as already small enough.
\section{Iterative compression}
\label{sec:iterative_compression}
In our method, we perform compression iteratively. We show that iterative approach helps to achieve higher compression ratio providing less accuracy drop. Our compression algorithm is defined as follows:

\begin{algorithm}
\caption{Iterative low-rank approximation}\label{alg:algorithm1}
\begin{algorithmic}[1]

\State $model \gets \text{load the model}$
\Statex
\Repeat
\State $accuracy \gets$ \Call{CheckAccuracy}{$model$}
\State $R_{i} \gets$ \Call{CheckInitialRanks}{$model$}
\State $R_{e} \gets$ \Call{VBMF}{$model$}
\State $R_{w} \gets$ \Call{RankWeakening}{$R_{e}$}
\Statex
\If {$\Call{GetConvolutional}{model} > 0$}
\State $model_{d} \gets$ \Call{HOSVD}{$model$, $R_{w}$}
\EndIf
\Statex
\If {$\Call{GetFullyConnected}{model} > 0$}
\State $model_{d} \gets$ \Call{SVD}{$model$, $R_{w}$}
\EndIf
\Statex
\State $model_{d} \gets$ \Call{FineTune}{$model_{d}$}
\State $accuracy_{d} \gets$ \Call{CheckAccuracy}{$model_{d}$}
\State $model \gets model_{d}$
\Until{$(accuracy - accuracy_{d}) < 1\%$}
\Statex
\State \Call{SaveModel}{$model$}
\end{algorithmic}
\end{algorithm}

Compressed model looses about 30\% of accuracy after applying decomposition but fine-tuning helps to recover original accuracy. It takes from 5 to 15 epochs for each iteration. If the model contains layers other than Convolutional and Fully connected, algorithm should skip most of them but for such layers as Batch normalization, model may require pre-processing to eliminate them.

\pagebreak
\section{Experimental results}
In this section we evaluate our method using three deep convolutional neural networks for classification and object detection:  AlexNet, VGG-16, YOLOv2 and Tiny YOLO. The first two networks were trained, fine-tuned and evaluated on \mbox{ImageNet \cite{deng2009imagenet}} dataset. The last one was trained and fine-tuned on PASCAL VOC2007 and PASCAL VOC2012 datasets and evaluated on PASCAL VOC2007 dataset. All tests were performed on CPUs of two different series and on GPU: Intel Core i5-7600K, Intel Core i7-7700K and NVIDIA GeForce GTX 1080 Ti respectively. 

\subsection{Results and discussion}
\label{ssec:results_and_discussion}
To demonstrate difference between our approach and non-repetitive one, we performed compression using the global analytic VBMF as the only rank selector. Results are shown below:

\begin{table}[!htb]
\begin{center}
\begin{tabular}{lcccc}
\hline
Model & Size & CPU1 & CPU2 & GPU \\
\hline
AlexNet & $\times5.1$ & $\times4.76$ & $\times4.82$  & $\times2.30$ \\
VGG-16 & $\times1.2$ & $\times2.65$ & $\times2.63$ & $\times1.75$ \\
YOLOv2 & $\times1.7$ & $\times1.70$ & $\times1.72$ & $\times1.47$ \\
Tiny YOLOv2 & $\times2.11$ & $\times1.98$ & $\times2.03$ & $\times1.58$ \\
\hline
\end{tabular}
\end{center}
\caption{Results of one-time compression for AlexNet, VGG-16, YOLOv2 and Tiny YOLO.}
\label{table1}
\end{table}

\noindent
Results of our method are shown below:
\begin{table}[!htb]
\begin{center}
\begin{tabular}{lcccc}
\hline
Model & Size & CPU1 & CPU2 & GPU \\
\hline
AlexNet & $\times4.90$ & $\times4.73$ & $\times4.55$  & $\times2.11$ \\
VGG-16 & $\times1.51$ & $\times3.11$ & $\times3.23$ & $\times2.41$ \\
YOLOv2 & $\times2.13$ & $\times2.07$ & $\times2.16$ & $\times1.62$ \\
Tiny YOLOv2 & $\times2.30$ & $\times2.35$ & $\times2.28$ & $\times1.71$ \\
\hline
\end{tabular}
\end{center}
\caption{Results of iterative low-rank approximation for AlexNet, VGG-16, YOLOv2 and Tiny YOLO.}
\label{table2}
\end{table}

\noindent
Comparing the results of compression ration and speed up in two tables, we can say that iterative approach showed itself better for almost all networks. We can also see that the highest speed up is achieved on CPUs. Moreover, we can see that speed up for different processor series is almost the same.

\pagebreak
\noindent
Comparison of accuracy losses between iterative method and one-time compression is shown below:

\begin{table}[!htb]
\begin{center}
\begin{tabular}{lcc}
\hline
Model & Iterative & One-time \\
\hline
AlexNet & $-0.81\%$ & $-4.2\%$ \\
VGG-16 & $-0.15\%$ & $-2.8\%$ \\
YOLOv2 & $-0.19\%$ & $-3.1\%$ \\
Tiny YOLO & $-0.10\%$ & $-2.7\%$ \\
\hline
\end{tabular}
\end{center}
\caption{Accuracy loss after iterative compression and one-time compression.}
\end{table}

\noindent
To achieve these results, fine-tuning has been performed as long as error rate decreased for both iterative approach and one-time compression approach. For AlexNet and VGG we measured Top-5 accuracy and for YOLOv2 and Tiny YOLO we measured mAP.

\section{Conclusions}
\label{sec:conclusions}
In this paper we addressed the problem of compression of deep convolutional neural networks. We proposed an iterative algorithm that performs gradual noise reduction. Our method consists of four repetitive steps:  automated selection of extreme rank values, rank weakening, tensor decomposition according to weakened rank and fine-tuning. We tested our approach on three deep networks (AlexNet, VGG-16, YOLOv2 and Tiny YOLO) used for classification and object detection. Experimental results show that our iterative approach outperform non-repetitive ones in the compression ratio providing less accuracy drop. As future work we will investigate effect of combining 
our approach with hardware-dependent approaches such as quantization.

%\vfill
%\pagebreak

%\newpage
%\clearpage
\bibliographystyle{IEEEbib}
\bibliography{refs}

\end{document}